# User Interface Tools for Navigation in Conditional Probability Tables and Elicitation of Probabilities in Bayesian Networks

Haiqin Wang & Marek J. Druzdzel
Decision Systems Laboratory
School of Information Sciences and Intelligent Systems Program
University of Pittsburgh, Pittsburgh, PA 15260
{whq,marek}@sis.pitt.edu

## Abstract

Elicitation of probabilities is one of the most laborious tasks in building decision-theoretic models, and one that has so far received only moderate attention in decision-theoretic systems. We propose a set of user interface tools for graphical probabilistic models, focusing on two aspects of probability elicitation: (1) navigation through conditional probability tables and (2) interactive graphical assessment of discrete probability distributions. We propose two new graphical views that aid navigation in very large conditional probability tables: the CPTREE (Conditional Probability Tree) and the sCPT (shrinkable Conditional Probability Table). Based on what is known about graphical presentation of quantitative data to humans, we offer several useful enhancements to probability wheel and bar graph, including different chart styles and options that can be adapted to user preferences and needs. We present the results of a simple usability study that proves the value of the proposed tools.

## 1  INTRODUCTION

Elicitation of numerical parameters is one of the most laborious tasks in building probabilistic models. The foremost problem is the large number of parameters required to fully quantify a model. For example, in a Bayesian network model created for diagnosis of liver disorders, HEPAR (Oniśko, Druzdzel, & Wasyluk 1998), there are 94 variables: the *Disorder* variable with 16 outcomes, and 93 feature variables. Full quantification of the HEPAR network required over 3,700 numerical parameters. In most real problem domains, elicitation of numerical parameters is a dominant task in probabilistic modeling (e.g., Heckerman, Horvitz, & Nathwani, 1992; Henrion 1989; Druzdzel & van der Gaag 1995).

Human judgement is prone to systematic errors (biases) that can be invoked by a variety of factors (Kahneman, Slovic, & Tversky 1982). Elicitation of probabilities, if not performed carefully, can result in poor quality estimates. Behavioral decision theorists have proposed several elicitation approaches that minimize the risk of bias. However, these methods tend to be cumbersome and often infeasible for models that include more than a few variables because of the large number of elicitations required. Decision analytic practice is usually based on methods that require less effort and still protect subjective assessments from common biases.

A major obstacle to effective probability elicitation in Bayesian networks is navigation in large conditional probability tables (CPTs). In a CPT, a conditional probability distribution over a variable is required for each combination of values of its parents. The total size of the conditional probability matrix is exponential in the number of parents. For example, the CPT of a binary variable with $n$ binary parents requires $2^{n+1}$ parameters. For a sufficiently large $n$, the $2^{n+1}$ numbers will not fit on the screen and the user will have to spend a considerable effort in navigating through them. The problem of navigation in conditional probability tables has not really surfaced in the field of decision analysis, as the size of typical decision-analytic models has been limited to a handful of variables. Bayesian networks, however, quickly reach the size of tens or hundreds of variables. It is not uncommon to see a variable with as many as ten parents, which, even if each parent is binary, results in CPTs consisting of thousands of elements. Existing software packages implementing Bayesian networks and influence diagrams have coped with the problem in various ways, few of which seems to follow established principles of good human-computer interface design. Users have to scroll back and forth to locate a particular item in a table or a list or have to manually give the combination of parent states in a combo box. Separate tables, applied in some solutions, require significant mental effort when users shift from one view to another.

While the problem of graphical elicitation of probabilities is easier to cope with, our investigation into the existing implementations has also shown a lot of room for improvement. The only graphical tool for probability elicitation implemented seems to be the probability wheel, which visualizes discrete probability distributions in a manipulable pie-chart graph. However, probability wheel has some problems and may sometimes be not the best tool for graphical elicitation of probabilities. A pie chart is known to make the judgement of part-to-part proportion difficult and is often inferior to a bar graph. Also, the labeling style applied and the overall design of interaction with the user is far from ideal in a typical implementation.

In this paper, we propose a set of tools that aim at improving navigation through large CPTs and at improving interactive assessment of discrete conditional probability distributions. We developed two new navigation tools: the CPTREE (conditional probability tree) and the sCPT (shrinkable conditional probability table). The CPTREE is a tree view of a CPT with a shrinkable structure for any of the conditioning parents. The sCPT is a table view of a CPT that allows to shrink any dimension of the table. Both CPTREE and sCPT allow a user to efficiently navigate through CPTs and, in particular, to quickly locate any combination of states of conditioning parents. We enhanced the probability wheel by providing alternative chart styles, bar graphs and pie charts, to support different kinds of proportion judgement. Our pie chart and bar graph support locking functions for those probabilities that have been elicited. Two labeling styles are provided: text and percentage. We use center-surround labels for the pie charts. While both tools are viable alternatives for probability elicitation, pie chart supports more accurate assessment of part-to-whole proportion whereas bar graph performs better for part-to-part proportion judgements. Both tools support context-specific independence and allow for elicitation of several distributions at a time, if these are identical.

The remainder of this paper is organized as follows. Section 2 offers a brief introduction to Bayesian networks and CPTs. Section 3 describes existing approaches to navigation in CPTs and existing implementations of graphical probability elicitation tools. Section 4 describes the CPTREE and sCPT and discusses the enhancements to the graphical elicitation tools. We report some findings from an empirical study based on the developed tools in Section 5.

## 2 BAYESIAN NETWORKS AND CONDITIONAL PROBABILITY TABLES

Bayesian networks (Pearl 1988) (also called *belief networks*, or *causal networks*) are a modeling tool that allows for an explicit representation of random variables and probabilistic interactions among them. Formally, they are directed acyclic graphs in which nodes represent random variables and arcs represent direct probabilistic influences among them.

An example of a Bayesian network is given in Figure 1. In this network, node *Disorder* has three bi-nary parents: *Alcoholism*, *Hepatotoxic medications*, and *Gallstones*, each of which is a causal factor contributing to each of six possible liver disorders. There are five symptoms of the modeled diseases, *Fatigue*, *Proteins*, *Ascites*, *Phosphatase* and *Jaundice*. Each node that has no incoming arcs (*Gallstones*, *Alcoholism*, and *Hepatotoxic medications*) is described by a prior probability distribution over its outcomes (e.g., fractions of patients who are alcoholic and not alcoholic in case of the binary node *Alcoholism*). In case of a node with incoming arcs, the network encodes its probability conditional on the outcomes of its direct predecessors (parents). For example, node *Disorder* is described by a probability distribution conditional on variables *Alcoholism*, *Hepatotoxic medications*, and *Gallstones*. Since all probabilities in our network are discrete, the conditional probability distribution is encoded as a conditional probability table (CPT). The CPT for the node *Disorder* is a four-dimensional table, where the first three dimensions are indexed by *Disorder*'s parents, *Alcoholism*, *Hepatotoxic medications*, and *Gallstones* and the fourth dimension is indexed by the outcomes of *Disorder*. The elements of this CPT are conditional probabilities of outcomes of *Disorder* given a combination of outcomes of *Alcoholism*, *Hepatotoxic medications*, and *Gallstones*.

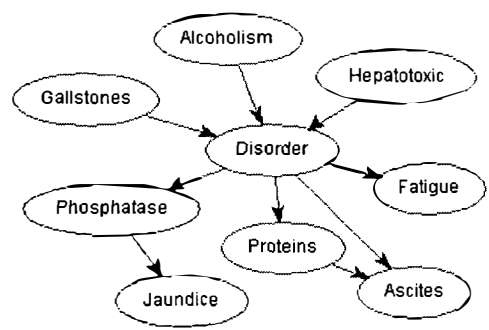

Figure 1: A Simplified Fragment of the HEPAR Network.

## 3 EXISTING GRAPHICAL TOOLS

Most of the existing probabilistic modeling systems provide graphical interface for navigation in conditional probability tables. Some of them supply a probability wheel as a graphical tool for subjective probability elicitation. An annotated list of these systems (including GENIE and each of the systems that we are referring to in this paper, DATA, DPL, ERGO, HUGIN, MSBN, and NETICA) along with links to their sites, where demonstration versions can be examined, is available on a web page maintained by the second author for the INFORMS' Society for Decision Analysis at http://www.sis.pitt.edu/~dsl/da-software.html. In this section, we analyze critically existing (at the time of this publication) graphical elicitation and navigation tools.

### 3.1 NAVIGATION IN CONDITIONAL PROBABILITY TABLES

There are several existing ways of dealing with the problem of navigation in conditional probability tables.

In a flat table, the solution adopted in GENIE 1.0 and HUGIN (Figure 2), the header cells indicate parent states and the numerical cells display the conditional probability distributions. The parent states are organized in a hierarchical structure that labels the conditional probability distributions. A table is a natural view for multi-dimensional data and, when it fits on the screen, it is fairly easy to explore. However, when a table is larger than the available screen area (this happens very often given the exponential growth of CPTs), users have to scroll back and forth to locate a particular conditional probability distribution. Watts (1994) observed that users of very large spreadsheets were often lost and ended up creating paper maps to guide them in navigation.

Figure 2: HUGIN's Probability Definition Table for Node *Disorder* in the HEPAR Network.

Another approach is using a list, a solution applied in NETICA (Figure 3). NETICA's navigation screen consists of a list of all possible combinations of parent states. The list of conditional probability distributions associated with each parent combination is shown next to the list of parent outcomes. Both lists are viewed by scrolling. If there are more items than those which can be shown in a list view, a scroll bar is provided for users to look for the hidden items. In Figure 3, two parents of node *Disorder* are shown in the parent list. The third one is hidden. The list in NETICA can be viewed as a transposed matrix of the table in GENIE and HUGIN. But the hierarchical structure is not clear in the list. Users are required to manually traverse the hierarchy to determine its structure. Generally, lists are capable of providing detailed content information but are poor at presenting structural information. A great deal of effort is needed on the part of the user to achieve a mental model of the structure in large hierarchies.

Figure 3: NETICA's Probability Definition Table for Node *Disorder* in the HEPAR Network.

Yet another solution is based on combo boxes, applied in MSBN (Figure 4). There is one combo box for each parent and it is used to select an outcome for that parent. Only one column of the CPT is visible at a time. In order to select a column, the expert has to assign values to all of the parents manually. When there are many parents, there is a danger that the user will forget to assign some of these combinations and, effectively, leave some of the probabilities unspecified.

Separate tables for parent combinations and conditional probabilities are yet another solution. ERGO uses two separate tables, one for a parent list and the other for a probability editor (Figure 5). When editing conditional probabilities for a node, the last parent is displayed in the probability editor table, and all other parents are displayed in the parent list. Separate tables show the conditional probability distribution for one combination of values of parents at a time, oc-cupying relatively small screen space. However, it is important to recognize that shifts from one table to another can be cognitively costly (Woods 1984).

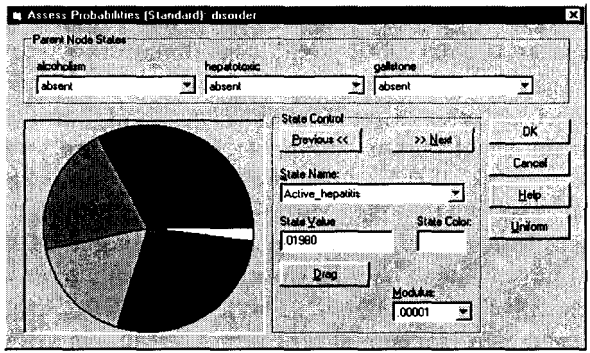

Figure 4: MSBN's Probability Definition Table for Node *Disorder* in the HEPAR Network.

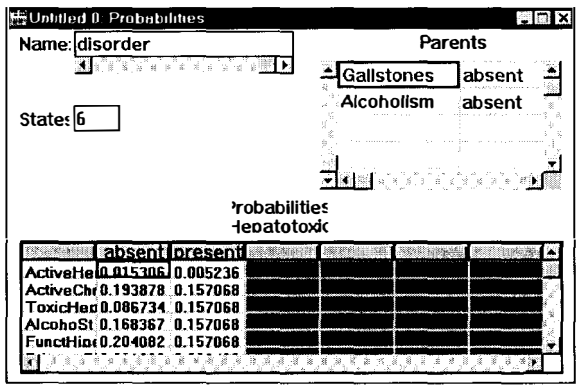

Figure 5: ERGO's Probability Definition Table for Node *Disorder* in the HEPAR Network.

A probability tree is a natural and familiar metaphor for the organization of conditional probability information. DPL provides a probability tree showing all of the possible combinations of parent outcomes (Figure 6). In DPL, the tree is always completely expanded and the entire tree appears in the available display space. The program shrinks the tree as needed to fit it on the screen. There is no zooming function for a clear view. The tree view provides a visual hierarchy of the context for specification of conditional probabilities. However, a completely expanded tree in a restricted display space becomes quickly unreadable. It is almost impossible to navigate in the tree view without remembering the order of parents and their outcomes.

### 3.2 ELICITATION OF PROBABILITIES

Probability wheel (Spetzler & Staël von Hostein 1975; Merkhofer 1987) is probably the oldest and the most popular graphical tool for probability elicitation. It consists of a wheel with two adjustable sectors (traditionally colored red and blue) and a fixed pointer in the center. When spun, the wheel will finally stop with the pointer either in red or blue sector. The probability that the wheel will stop in the red sector is proportional to the sector size. The relative size of the two sectors can be adjusted until the expert judges that the event under consideration is equally likely as the event of the wheel stopping in the red region. In computer systems (e.g., DATA, DPL and MSBN), it is usually implemented as a pie chart. The pie chart is partitioned into several sectors representing each of the outcomes of the variable. The area of each sector is proportional to the probability of the corresponding outcome. The user can shrink or expand the proportion of each area by dragging its edge to the desired position.

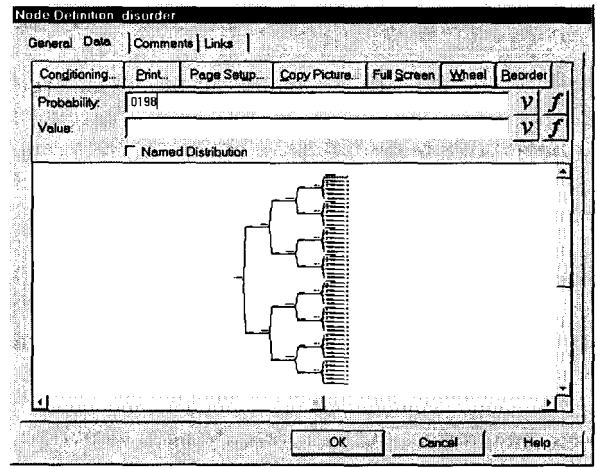

Figure 6: DPL's Probability Definition Table for Node *Disorder* in the HEPAR Network.

While the probability wheel is a useful tool, it has several disadvantages. Probability elicitation involves complex perceptual processes that include judgements of proportions, comparisons, and changes. Graphical tools help experts to estimate proportions, and to dynamically change the sizes of component parts in the graph until the sizes reflect personal beliefs of the experts. When eliciting subjective probabilities, some experts find it difficult to judge a part-to-whole proportion. They often use a larger value as reference point and compare smaller values with it for a part-to-part judgement. Although empirical studies have demonstrated that pie charts lead to a higher accuracy in part-to-whole judgement of proportion, they have shown inferiority of pie charts to bar graphs in part-to-part comparison and change perception (Cleveland & McGill 1984; Simkin & Hastie 1987; Hollands & Spence 1992; 1998). A pie chart has the additional disadvantage of being too fragmentary when partitioned into many sectors. Preece *et al.* (1994) recommended that a pie chart should be used only when there are fewer than five sectors.

Lack of user control is another problem with the existing implementations of probability wheel. Since total probability is always equal to one, a specific change in probability of one outcome results in proportional changes in the probability of the remaining outcomes. The proportion of the remaining outcomes usually stays the same. However, this automatic adjustment of probabilities is frustrating when an expert just wants to modify some of the numbers and keep other numbers unchanged. This happens, for example, when the expert accepts some probabilities encoded and only wants to graphically modify remaining probabilities. It seems necessary to let the expert be in control of when and to which probabilities the automatic changes apply.

Besides, the legend annotation of a pie chart requires a mapping procedure to recognize which sector represents which outcome of the variable. When a variable has many outcomes, it becomes difficult for the user to search in a long list for those mappings between outcomes and sectors. The coordination of human focal attention and orienting perceptual functions such as peripheral vision supports the process of knowing where to look, and when (Rabbitt 1984). Woods [1984] suggested the use of a center-surround technique, which is an annotation style that labels wedges around sectors of the pie. A direct label is provided for each sector, thus reducing the mental workload of the users.

## 4 GRAPHICAL TOOLS DEVELOPED

In this section, we describe the graphical tools that we have developed for the purpose of elicitation of probabilities in conditional probability tables (CPTs).

### 4.1 NAVIGATION IN CPT'S

As we discussed in the previous section, the plain form of a CPT is hard to navigate due to the exponential growth of its size. In order to address this problem, we adopted the tree metaphor for hierarchical visual representations and developed two browsing tools: the CPTREE (conditional probability tree) and the sCPT (shrinkable conditional probability table).

#### 4.1.1 Conditional Probability Tree

The CPTREE (Figure 7) is a tree view of a node's CPT. In a CPTREE, every parent variable is represented by two levels of nodes, the name level and the outcome level. The name level is comprised of a single node indicating the variable's name. The outcome level includes nodes for all possible outcomes of the corresponding variable. The name node always appears as the parent of the outcome nodes for the same variable. Each name node is a child of an outcome node of the previous parent variable. The root of the CPTREE is the name node of the first parent variable. The path from root to a leaf specifies a combination assignment to values of parents. On the right-hand side of the tree is a table in which each row is associated with a branch in the tree. The table defines the probabilities conditional on the context specified by the branches.

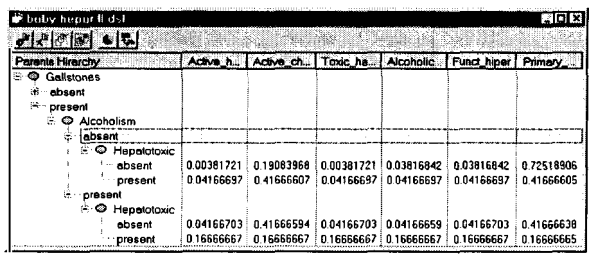

Figure 7: CPTREE of Node *Disorder* in a Simplified Version of the HEPAR Network.

With shrinking and expanding function, an expert can quickly go to the branches of interest while collapsing others in order to optimize screen use. A click on the corresponding toolbar icon will bring up a probability wheel for the probability distribution conditioned on the selected combinations of the parent assignments for the current node represented by the CPTREE. A combination of parent assignments is specified by a path from the root to a leaf in the CPTREE. If a leaf node is selected, the conditioning context is given by all of the parent assignments along the path. If an internal node in the CPTREE is selected, the context is given by a partial combination of parent assignments. Only those parents that are between the root of the tree and the selected node will count. For example, in Figure 7, when the tree node *absent* under *Alcoholism* is selected, the selected branch specifies the context of *Gallstones* = *present* ∧ *Alcoholism* = *absent*. The state of *Hepatotoxic* is irrelevant. In other words, the probability distribution over *Disorder* is independent of the state of *Hepatotoxic*. The selected branch defines a context-specific independence (also called asymmetric independence) relationship (Shimony 1993; Boutilier *et al.* 1996; Geiger & Heckerman 1996) between the current variable, *Disorder*, and its parent, *Hepatotoxic medications*.

Our design of the navigation interface allows the user to dynamically change the order of the parents in the navigation windows. Many times the users of GeNIe 1.0 found the order of node parents counterintuitive because it did not follow the temporal or causal order. Changing the order of parents as the user desires allows the user to compose the most natural order of conditioning events. Secondly, it facilitates easy encoding of context-specific independence.

Multiple selection of branches is also supported. By selecting multiple branches and then triggering graphical elicitation through the probability wheel, experts can give their assessment for those conditional probabilities that are numerically identical but different in conditions. In Figure 7, the conditional probabilities under the context of $Gallstones = present \wedge Alcoholism = absent \wedge Hepatotoxic = present$, and the conditional probabilities under the context of $Gallstones = present \wedge Alcoholism = present \wedge Hepatotoxic = absent$ can be estimated at the same time by selecting both of the corresponding branches. Using this multiple assignment, experts can save a lot of duplicate input, which often happens in flat CPTs of current graphical probabilistic modeling development environments.

#### 4.1.2 Shrinkable Conditional Probability Table

The sCPT (Figure 8) includes virtually all of the functions implemented in the CPTree. Double-clicking on a header item triggers the shrinking or expanding of the columns that it covers. We can view the sCPT as a tree-structured conditional probability table. All the columns in the covered range of a header item constitute its children items. A branch can be traced from the first header row through its covered range. With the aid of probability tools, experts can assign the same probability values to multiple groups under distinct branches.

Figure 8: A Shrinkable CPT of Node *Disorder* with the *Gallstones=absent* Branch Shrunk.

Compared to the CPTree, the sCPT has a higher data density, which is a desired property of graphical displays of quantitative data, defined as the ratio of the amount of data displayed to the area of the graphic (Tufte 1998). In the CPTree, a considerable amount of screen area is consumed at the expense of displaying the dependence context for conditional probabilities. This results in the difficulty of the CPTree to represent a node with a large number of parents. However, for some users, the CPTree may visualize the structure of conditional dependence more intuitively.

### 4.2 PROBABILITY ASSESSMENT TOOLS

We have designed a graphical tool for elicitation of discrete probability distributions that implements two chart styles: pie charts and bar graphs. When the user selects the assessment tool within the navigation window, the tool is presented in a separate pane of the splitter window of the navigation tool.

Our pie chart (Figure 9) combines easy user interaction with intuitive illustration. To change a probability of an outcome of a variable, the user drags the handle of the corresponding sector in the pie to its new position. During the dragging process, the pie is redrawn, showing the new partition resulting from the probability changes. When one probability is being changed, the remaining probabilities are automatically adjusted proportionally. If the user wants to keep the probabilities of some events intact, she can simply click the right mouse button on the sectors corresponding to these events to lock them before beginning the dragging process. A right click on a locked sector unlocks it. A locked sector of the pie is shaded out and drawn slightly outside the pie, visually communicating the idea that this part of pie is cut off and cannot be changed. In Figure 9, two outcomes of *Disorder*: *Toxic_hepat* and *Active_chron* are locked and shaded out of the whole pie.

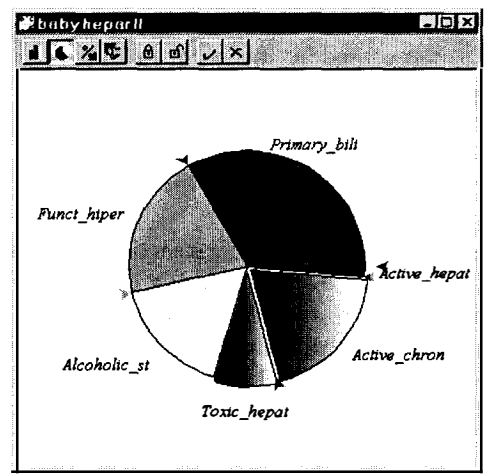

Figure 9: Pie-chart-styled Probability Wheel for Node *Disorder* in the HEPAR Network with two Locked Sectors: *Toxic_hepat* and *Active_chron*.

Our bar graph (Figure 10) provides a similar functionality. The user can adjust the length of a bar by dragging the handle at its end horizontally to a new position. The unlocked bars are changed proportionally, while the locked bars remain unchanged during the adjusting process. All locked bars are shaded in their vertical color gradients. Figure 10 shows a probability elicitation tool styled as a bar graph with probability scale appearing on the bottom of the graph. *Toxic_hepat* and *Active_chron* are locked and shown in their vertical color gradients.

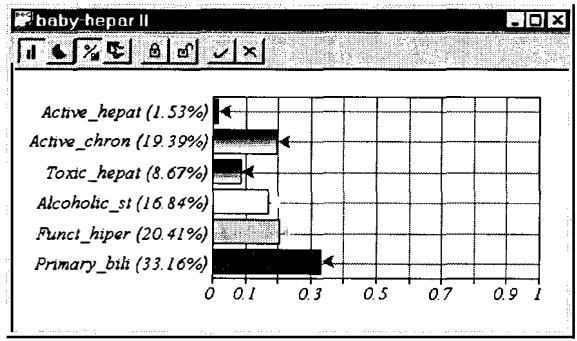

Figure 10: Bar-graph-styled Probability Tool for Node *Disorder* in the HEPAR Network with two Locked Sectors: *Toxic_hepat* and *Active_chron*. Note Optional Percentage Labeling.

We provided two labeling options for both chart styles. One is simple text of the outcome name as shown in Figure 9. The other is the name plus its probability as shown in Figure 10. Text labeling eliminates the interference of numbers and leads to a qualitative estimation from experts. Percentage labeling allows experts to see the exact numerical parameters corresponding to the manipulated graph components.

In addition, we use center-surround labels for the pie chart. Labels are positioned outside the pie near their corresponding sectors. This supports the user's perceptual process of knowing where to look and when, and reduces the mental workload of mapping labels from legend annotation to the corresponding sectors in the pie. When there are overlaps for two adjacent labels, only the last one is displayed, the next label is hidden.

Both charts are viable alternatives for probability elicitation tools, although they serve different purposes. Pie charts are more natural to show the relative distribution of data among sectors that make up a whole. Generally they support more accurate assessment of part-to-whole proportion. But our bar graphs are supplemented with a scale ranging from zero to one, which also facilitates assessment of part-to-whole proportion. On the other hand, people sometimes do not have a clear idea about what proportion a part takes up in a whole. But they often can give a proportion of one part to another by comparing them. According to the ranking of human perception identified by Cleveland and McGill (1984), people usually produce faster and more accurate judgements when comparing position and length than when comparing angle or area. In addition, bar graph has an advantage over pie chart in the perception of change. People can easily capture small changes in a bar graph. Thus, a bar graph can be expected to allow for a better performance in probability estimation based on part-to-part proportion.

## 5 IMPLEMENTATION AND EMPIRICAL VERIFICATION

We have implemented the tools described in this paper in a forthcoming GENIE 2.0, an environment for building graphical decision-theoretic models, under development at the Decision Systems Laboratory, University of Pittsburgh. User interface has received a considerable attention in GENIE. We believe that GENIE's growing popularity (over 2,200 users of GENIE 1.0 as of May 2000) can be in part attributed to our attention to detail and the resulting powerful, yet pleasant interface. One of our objectives is to enhance GENIE's interface so that it becomes natural and easy to use for both experts and novices. We believe that it is not the speed of inference but rather the quality of the user interface that will increase the popularity of decision-theoretic methods. In our experience, reasoning in most practical models is sufficiently fast. The current bottleneck is in building models. Therefore, techniques that facilitate model building and intuitive interaction with the system are worth pursuing, even if they are cumbersome to implement in software.

All GENIE 2.0 windows are fully resizable. Users can always see a larger view of a CPTREE or a probability wheel by enlarging an appropriate window. The size of the pie chart or bar graph is adjusted automatically to fit in the newly resized window. A relevant detail of our implementation is that GENIE's models are always syntactically correct at any stage of model development. A newly added node, in particular, has by default two outcomes, *State0* and *State1*, that are uniformly distributed. Any additional operation preserves this correctness. A negative side-effect of this is that the program does not have a clear way of showing which probabilities have been elicited and which have not.

While there have been other studies testing graphical representation of numerical data (e.g., Feldman-Stewart *et al.*, 2000; Sparrow, 1989), none of them focused on elicitation of probabilities. We tested empiri-cally the two graphical probability elicitation methods, pie chart and bar graph, on a task involving elicitation of conditional probability distributions (Wang, Dash, & Druzdzel 2000). The results of our test have shown statistically significant differences in both speed and accuracy between each of the two methods and direct elicitation of numerical probabilities. Even though graphical probability elicitation methods were both faster than direct elicitation of numerical probabilities, bar graph was a clear winner in terms of both accuracy and speed (11% more accurate and 41% faster than direct elicitation and 3% more accurate and 35% faster than the pie chart). Space constraint does not allow us to report the details of that study in the current paper.

Qualitative questionnaire conducted at the conclusion of the study has shown that our subject valued highly the availability of navigational tools and in majority of cases preferred graphical elicitation to direct numerical input of probabilities (there were some exceptions). As far as preference between the two graphical modes is concerned, we noticed that it varied between subjects, suggesting that a good tool should provide a variety of methods that can adjust to individual user preferences.

## 6 CONCLUDING REMARKS

The tools proposed in this paper enhance greatly user navigation in CPTs during the process of model building and help to improve both the quality and speed of elicitation. Also, the flexible navigation and visualization of probability distributions help to detect unspecified probabilities and inconsistency in responses. Combined, these tools provide a pleasant and powerful visual environment in which experts can give their qualitative estimates of numerical probabilities.

While the tools that we have designed and implemented may be applicable to other graphical probabilistic structures, such as chain graphs, we focused on Bayesian networks. Obviously, the methods are also applicable to chance nodes in influence diagrams (Howard & Matheson 1984). We plan extending these tools to utility nodes and chance nodes described by canonical probabilistic interactions, such as Noisy-OR or Noisy-AND nodes. A useful enhancement to the bar graph tool will be marking it with user-defined probability scales, such as verbal probabilities, that will for some users enhance the elicitation process even further.

We did not use 3-D displays, even though extra dimensions are often decorative and attractive. Some experiments (Spence 1990; Carswell, Frankenberger, & Bernhard 1991; Siegrist 1996) evaluated 3-D graphs in a perception task of relative magnitude estimation. The results did not show an advantage of 3-D displays in accuracy and speed. The performance of 3-D displays depends on the graphs and tasks. Compared to their 2-D counterparts on relative magnitude estimation, 3-D pie charts result in lower accuracy, and the 3-D bar graphs require a longer elicitation time.

One limitation of the current version of the assessment tools is lack of support for elicitation of very small probabilities. Due to the screen resolution restriction and sensitivity of mouse movement, it is hard to capture very small changes of mouse position. Therefore, it is impossible to distinguish between low probabilities such as 0.000001 and 0.00001, even though they are orders of magnitude apart. Such values have to be entered manually. Using longer bars or bigger pies can improve the assessment accuracy of very low probabilities, but this consumes more screen resources. A good solution applied by others is the log-scale (López Gómez 1990).

## Acknowledgments

The research was supported by the Air Force Office of Scientific Research grants F49620-97-1-0225 and F49620-00-1-0112 and by the National Science Foundation Faculty Early Career Development (CAREER) grant IRI-9624629. While we are sorely responsible for any errors in the paper, we would like to thank Javier Díez and anonymous reviewers for valuable feedback. We are grateful to Michael Lewis and Stephen Hirtle for their guidance through the Human-Computer Interaction literature and valuable suggestions. Our colleagues in the Decision Systems Laboratory contributed to the design of the tools described in this paper with their suggestions.